\pdfoutput=1

\documentclass[11pt]{article}

\usepackage{acl}

\usepackage{times}
\usepackage{latexsym}
\usepackage{mathtools}
\usepackage{amssymb}
\usepackage{times}
\usepackage{latexsym}
\usepackage{booktabs}
\usepackage{soul}
\usepackage{xcolor}
\usepackage{subcaption}
\usepackage{url}
\usepackage{graphicx}
\usepackage{multirow}
\usepackage{amssymb}
\usepackage{pifont}

\usepackage[T1]{fontenc}

\usepackage[utf8]{inputenc}

\usepackage{microtype}

\usepackage{siunitx}
\title{Biomedical Language Models are Robust to Sub-optimal Tokenization}

\author{Bernal Jim\'{e}nez Guti\'{e}rrez \quad Huan Sun \quad Yu Su
\\The Ohio State University
\\\small{\texttt{\{jimenezgutierrez.1,sun.397,su.809\}@osu.edu}}
}

\begin{document}
\maketitle
\begin{abstract}

As opposed to general English, many concepts in biomedical terminology have been designed in recent history by biomedical professionals with the goal of being precise and concise. 
This is often achieved by concatenating meaningful biomedical morphemes to create new semantic units. 
Nevertheless, most modern biomedical language models (LMs) are pre-trained using standard domain-specific tokenizers derived from large scale biomedical corpus statistics without explicitly leveraging the agglutinating nature of biomedical language. 
In this work, we first find that standard open-domain and biomedical tokenizers are largely unable to segment biomedical terms into meaningful components. 
Therefore, we hypothesize that using a tokenizer which segments biomedical terminology more accurately would enable biomedical LMs to improve their performance on downstream biomedical NLP tasks, especially ones which involve biomedical terms directly such as named entity recognition (NER) and entity linking. 
Surprisingly, we find that pre-training a biomedical LM using a more accurate biomedical tokenizer does not improve the entity representation quality of a language model as measured by several intrinsic and extrinsic measures such as masked language modeling prediction (MLM) accuracy as well as NER and entity linking performance. 
These quantitative findings, along with a case study which explores entity representation quality more directly, suggest that the biomedical pre-training process is quite robust to instances of sub-optimal tokenization.\footnote{Our code and pre-trained models are publicly available at \url{https://github.com/OSU-NLP-Group/Bio-Tokenization}}
\end{abstract}

\section{Introduction}
In order to communicate complex concepts precisely and efficiently, biomedical terminology has been designed by researchers and medical professionals by combining existing meaningful morphemes to create new concepts. Many biomedical terms use general rules to combine meaningful morphemes taken from Greek and Latin \cite{medical_term}. For example, these morphemes often have vowels that can be omitted such as the `-o-' in `nephro' from Table \ref{tab:tokenization-examples}. This `-o-' acts as a joint-stem to connect two consonantal roots (e.g. `nephr-' + `-o-' + `-pathy' = `nephropathy'), but the `-o-' is often dropped when connecting to a vowel-stem (e.g. `nephr-' + `-ectomy' = `nephrectomy', instead of `nephr-o-ectomy'). Students in biomedical fields often learn the meaning of these elements as well as the word formation rules to be able to infer the meaning of unfamiliar words and recall complex terms more easily.\footnote{The existence of popular books such as \citet{Collins2007-go} emphasize the importance of understanding biomedical terminology design for medical professionals.}

\begin{table}[!t]
\centering
\resizebox{\columnwidth}{!}{%
\begin{tabular}{@{}ccc@{}}
\toprule
\textbf{\begin{tabular}[c]{@{}c@{}}Ideal \\ Tokenization\end{tabular}}                    & \textbf{\begin{tabular}[c]{@{}c@{}}BERT\\ Tokenization\end{tabular}} & \textbf{\begin{tabular}[c]{@{}c@{}}PubMedBERT\\ Tokenization\end{tabular}} \\ \midrule
\begin{tabular}[c]{@{}c@{}}nephr-o-pathy\end{tabular}            & ne-ph-rop-athy                                                       & nephropathy                                                                \\
\begin{tabular}[c]{@{}c@{}}nephr-ectomy\end{tabular}             & ne-ph-re-ct-omy                                                      & nephrectomy                                                                \\
\begin{tabular}[c]{@{}c@{}}nephr-o-blastoma\end{tabular}      & ne-ph-ro-bla-sto-ma                                                  & nephr-oblastoma                                                            \\
\begin{tabular}[c]{@{}c@{}}nephr-o-calcin-osis\end{tabular} & ne-ph-ro-cal-cino-sis                                                & nephr-ocalcin-osis                                                         \\ \bottomrule
\end{tabular}%
}
\caption{Sub-optimally tokenized biomedical terms containing the `nephro' morpheme illustrate the limitations of current tokenization methods.}
\vspace{-5pt}
\label{tab:tokenization-examples}
\end{table}

Even though the agglutinating nature of biomedical terminology is well known, none of the existing pre-trained language models consider this information explicitly when building their tokenizers. As shown in Table \ref{tab:tokenization-examples}, frequent words such as `nephropathy' and `nephrectomy' are tokenized by BERT \cite{devlin-etal-2019-bert} into meaningless subwords (`ne-ph-rop-athy' and `ne-ph-re-ct-omy') while remaining as whole words for PubMedBERT \cite{pubmedbert}. For more infrequent but still important medical terms like `nephroblastomas' and `nephrocalcinosis', PubMedBERT encodes them as `nephr-oblastoma' and `nephr-ocalcin-osis'. We argue that there are more meaningful and efficient ways to tokenize both frequent and infrequent medical terms using meaningful morphemes like `nephr(o)' (of a kidney), `-pathy/-(o)sis' (disease), `-ectomy' (surgical procedure), `calcin' (calcification) and `blastoma' (type of cancer) which could help models transfer signal directly into infrequent and even out-of-vocabulary terms. 

In this work, we first leverage large-scale morpheme segmentation datasets to more rigorously evaluate current tokenization methods both quantitative and qualitatively. Using the annotated morpheme segmentation dataset from the SIGMORPHON 2022 Shared Task \cite{batsuren-etal-2022-SIGMORPHON}, we are able to determine that current tokenizers such as BERT and the more biomedically relevant PubMedBERT are very poorly aligned with human judgments on morpheme segmentation, even when evaluating on biomedical terminology specifically.

Given that, although the PubMedBERT tokenizer exhibits low performance on biomedical morpheme segmentation, it shows some improvement over BERT's tokenizer due to its use of biomedical corpus statistics, we hypothesize that using a tokenizer that aligns more strongly with standard biomedical terminology construction for pre-training could achieve improved performance in downstream tasks. In order to verify this hypothesis, we create a new tokenizer, BioVocabBERT, which uses a vocabulary derived from combining a fine-tuned morpheme segmentation model with biomedical domain-knowledge from the Unified Medical Language System (UMLS) \cite{umls}, a large scale biomedical knowledge base. Subsequently, we leverage BioVocabBERT, which greatly outperforms the PubMedBERT tokenizer on biomedical morpheme segementation, to pre-train a biomedical language model by the same name and compare its performance with a replicated PubMedBERT model (to control for any potential differences in the pre-training process) on two downstream tasks: named entity recognition (NER) and entity linking.

Surprisingly, we find that the performance of our BioVocabBERT model is remarkably similar to the one obtained by our PubMedBERT replica throughout most datasets tested in fully supervised NER, low-resource NER and zero-shot entity linking. Small improvements arise in low-resource NER and zero-shot entity linking but results are inconsistent across datasets. Additionally, we examine the model's robustness to segmentation failures in a small scale case-study which suggests that even if the model's word embeddings are biased by tokenization errors, the model's parameters are able to overcome such failures quite successfully. Finally, we measure our models' language modeling accuracy by word frequencies and find a small word frequency trade-off whose exploration we leave for future work. Given these findings, we conclude that biomedical language model pre-training is quite robust to tokenization decisions which are not well aligned with human judgments, even when dealing with highly agglutinating biomedical terminology. 

\section{Related Work}

\subsection{Domain-Specific Pre-training}

Recent work on domain-specific language models has demonstrated fairly conclusively that using domain-specific data for pre-training significantly improves language model performance on in-domain downstream tasks. Many different such strategies have been proposed with varying degrees of in-domain vs out-of-domain pre-training data in fields such as biomedicine \cite{peng-etal-2020-empirical, Lee2019BioBERTAP, pubmedbert, el-boukkouri-etal-2022-train}, finance \cite{Wu2023BloombergGPTAL}, law \cite{chalkidis-etal-2020-legal}, scientific research \cite{maheshwari-etal-2021-scibert}, clinical practice \cite{alsentzer-etal-2019-publicly} and social media \cite{delucia-etal-2022-bernice}. For biomedical language models specifically, most work agrees that pre-training from scratch using an in-domain corpus, as done by \citet{pubmedbert}, leads to small but measurable performance improvements over other pre-training strategies.

Apart from introducing pre-training from scratch, \citet{pubmedbert} demonstrated the limitations of general domain tokenization for domain-specific pre-training by showing downstream improvements obtained from using a domain-specific tokenizer, created by standard tokenizer building algorithms such as WordPiece \cite{Schuster2012JapaneseAK} and BPE \cite{Gage1994ANA} on an in-domain corpus. In-domain tokenizers have also been shown to improve performance in other domains such as law \cite{chalkidis-etal-2020-legal} and more specific ones like cancer \cite{Zhou2022CancerBERTAC}. As a result, most recent biomedical LMs use tokenizers created from in-domain corpora statistics \cite{yasunaga-etal-2022-linkbert, Luo2022BioGPTGP}.

\subsection{Limits of Unsupervised Tokenization}

Even though domain-specific tokenizers have become widely used for biomedical LMs, they are still constructed using mainly unsupervised algorithms which leverage information theoretic metrics from large-scale corpora to create subword vocabularies. However, as reported in the SIGMORPHON 2022 Shared Task for morpheme segmentation \cite{batsuren-etal-2022-SIGMORPHON}, these methods align little with morphological human judgments. \citet{hofmann-etal-2021-superbizarre} explore how poor segmentation affects performance by injecting rule-based derivational morphology information into the tokenization process and showing improvements in word classification tasks, especially for low-frequency words. As far as we know, our work is one of the first to perform a similar morpheme segmentation analysis on biomedical tokenizers, even though biomedical terminology is highly agglutinating by design and should benefit from such analysis.

Furthermore, \citet{hofmann-etal-2020-dagobert} shows that introducing derivational morphology signal into BERT via fine-tuning improves its derivation generation capabilities, suggesting that performance of language models could be improved by adding morphologically relevant signal into their pre-training. Nevertheless, not much work apart from our current study has explored how introducing such signals could affect the pre-training process directly, especially not in biomedical language models.

\section{Supervised Morpheme Segmentation}

Recent work evaluating morphological segmentation at scale such as the SIGMORPHON 2022 Shared Task \cite{batsuren-etal-2022-SIGMORPHON} demonstrates the impressive performance of supervised methods compared to unsupervised methods like BPE \cite{Gage1994ANA} or Morfessor \cite{Creutz2005UnsupervisedMI}, even for languages with more limited annotated data than English. In the SIGMORPHON 2022 Shared Task, the organizers compile a large quantity of segmented morpheme data, over half a million English words obtained from Wiktionary and other sources using both hand-crafted and automated methods \cite{batsuren-etal-2021-morphynet}.

\subsection{Evaluating Biomedical Segmentation}\label{sec:biomedical_subset}

By comparing the SIGMORPHON dataset with words which appear frequently in the Unified Medical Language System (UMLS) \cite{umls}, a large scale biomedical knowledge base, we find that a small percentage (approximately ~10\%) of all annotated words are relevant biomedical terms. We therefore leverage this biomedical subset to evaluate the biomedical morpheme segmentation performance of several current tokenization methods. Due to the large difference in scale of the full dataset to the biomedical subset, we use the full SIGMORPHON dataset for training, including both general english and biomedical words. We use the same segmentation F1 score the SIGMORPHON Shared Task for evaluation. This score is calculated as the harmonic mean of \textit{precision}, the ratio of correctly predicted morphemes over all predicted morphemes, and \textit{recall},  the ratio of correctly predicted
morphemes over all gold-label units. For more information about these evaluation metrics, we refer the interested reader to Section 2.3 of \citet{batsuren-etal-2022-SIGMORPHON}.

\begin{table}[!t]
\centering
\small
\resizebox{0.43\textwidth}{!}{%
\begin{tabular}{@{}lccc@{}}
\toprule
                & \textbf{Train} & \textbf{Dev} & \textbf{Test} \\ \midrule
English Set & 458,692           & 57,371          & 57,755           \\
Biomedical Subset & 33,221          & 4,112          & 4,123           \\ \bottomrule
\end{tabular}%
}
\caption{Dataset statistics for the SIGMORPHON 2022 morpheme segmentation dataset and the biomedical dataset, as defined in \S\ref{sec:biomedical_subset}.}
\label{tab:my-table}
\vspace{-6pt}
\end{table}

\begin{table}[!h]
\centering
\small
\resizebox{0.4\textwidth}{!}{%
\begin{tabular}{@{}lc@{}}
\toprule
                   & \textbf{Segmentation F1} \\ \midrule
\textbf{BERT Tokenizer}           & 16.2                   \\
\textbf{PubMedBERT Tokenizer}     & 19.2                   \\
\midrule
\textbf{Fine-Tuned CANINE}      & 74.1                   \\
\midrule
\textbf{BioVocabBERT Tokenizer} & 48.5                   \\ \bottomrule
\end{tabular}%
}
\caption{Morpheme segmentation performance of baseline and novel tokenizers on the biomedical subset of the SIGMORPHON 2022 development set.}
\label{tab:segmentation-eval}
\vspace{-5pt}
\end{table}
 
As seen in Table \ref{tab:segmentation-eval}, both BERT and PubMedBERT achieve under 20\% segmentation F1 performance on the SIGMORPHON biomedical development subset. In order to understand why current tokenizers obtain such dramatically low segmentation accuracy, we analyze 50 instances of sub-optimal tokenization. Apart from words which are not segmented because they exist in the PubMedBERT vocabulary, most errors are split into three main categories 1) missing units, 2) compound units and 3) ambiguous connecting vowels. Table \ref{tab:seg_error_types} shows descriptions and examples of each type.

\begin{table}[!h]
\centering
\resizebox{\columnwidth}{!}{%
\begin{tabular}{@{}lll@{}}
\toprule
\textbf{\begin{tabular}[c]{@{}l@{}}\end{tabular}}    & \textbf{Description}                                                                                                & \textbf{Example}                                                                                                                      \\ \midrule
\textbf{\begin{tabular}[c]{@{}l@{}}Missing \\ Units\end{tabular}}     & \begin{tabular}[c]{@{}l@{}}Important biomedical \\morphemes missing\\from the vocabulary\end{tabular} & \begin{tabular}[c]{@{}l@{}}\textbf{onc-oneu-ral} \\ `onco' (cancer-related)\\  is not in the vocabulary\end{tabular}                        \\\midrule
\textbf{\begin{tabular}[c]{@{}l@{}}Compound \\ Units\end{tabular}}    & \begin{tabular}[c]{@{}l@{}}Splitting meaningful \\units while creating\\ meaningless ones\end{tabular}                  & \begin{tabular}[c]{@{}l@{}}\textbf{neuroprot-ectant} \\  `neuroprot' is meaningless\\ and splits the meaningful\\ morpheme `protect'\end{tabular} \\\midrule
\textbf{\begin{tabular}[c]{@{}l@{}}Connecting \\ Vowels\end{tabular}} & \begin{tabular}[c]{@{}l@{}}Vowels which connect \\two morphemes \\ (more ambiguous)\end{tabular}                   & \begin{tabular}[c]{@{}l@{}}\textbf{bronch-olith}\\ optimal segmentation \\ could split `o' from `lith'\end{tabular}                         \\ \bottomrule
\end{tabular}%
}
\caption{Sub-optimal segmentation types from the biomedical subset of SIGMORPHON 2022.}
\label{tab:seg_error_types}
\end{table}

\subsection{CANINE Fine-Tuning}

As opposed to sub-word tokenization, morpheme segmentation does not require sub-word components (morphemes) to map directly onto a word's characters. For instance, in Table \ref{tab:canine-supervised-tokenizer}, SIGMORPHON annotations transform the root `neur' into the word `neuron', introducing further flexibility and complexity to the task. In order to adapt morpheme segmentation annotations to standard tokenization, we design rule-based heuristics that map each morpheme onto a subset of characters in the original word. Due to this new formulation and the success of transformer based models on this shared task, we choose a character based language model named CANINE \cite{clark-etal-2022-canine} as the model to train for character tagging as morpheme segmentation. More formally, the segmentation task is re-framed as classifying each character into a B(egin) or I(nside) tag, where the B tag indicates the start of a new morpheme or token. 

\begin{table}[!h]
\centering
\resizebox{0.45\textwidth}{!}{%
\begin{tabular}{@{}ccc@{}}
\toprule
\textbf{Original Word} & \textbf{Segmentation}   & \textbf{BI Tags} \\ \midrule
onconeural          & onco \#\#neur \#\#al & BIIIBIIIBI     \\ \bottomrule
\end{tabular}%
}
\caption{Example of a biomedical term segmented into morphemes and reformulated into BI tags for CANINE fine-tuning.}
\label{tab:canine-supervised-tokenizer}
\end{table}

\begin{figure*}[!t]
     \centering
     \includegraphics[width=\textwidth]{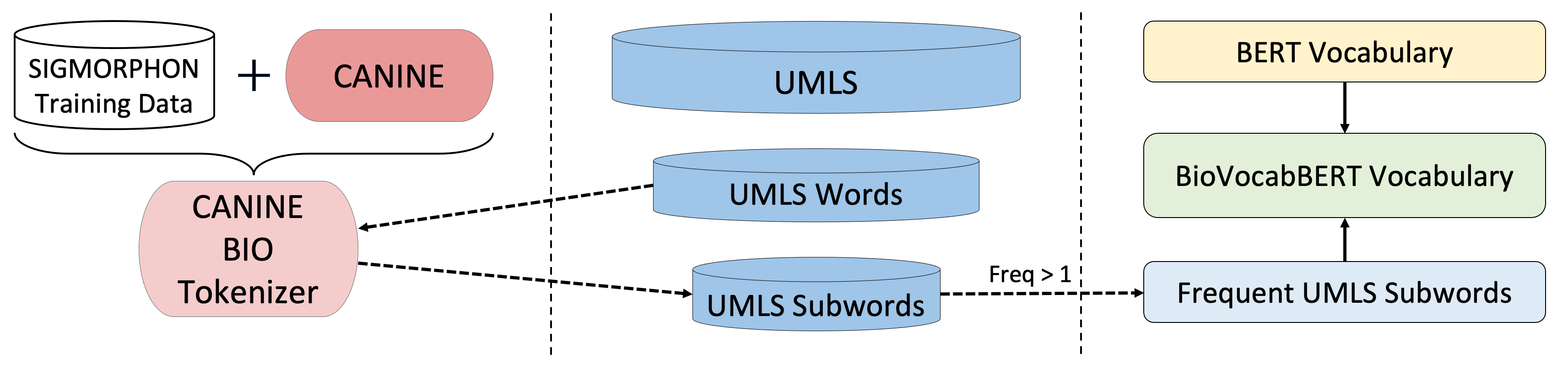}
     \caption{Overall process for creating our vocabulary for BioVocabBERT's tokenizer. We first train a CANINE model on the SIGMORPHON training set (left). We then use this segmentation model to segment all unique UMLS words (center). Finally, we combine all UMLS subwords with frequency greater than 1 with the original BERT vocabulary to make our BioVocabBERT vocabulary (right).}
     \label{fig:biovocabbert_constr}
\end{figure*}

After fine-tuning CANINE on the full SIGMORPHON 2022 training set to create a supervised tokenization system, as seen in Figure \ref{fig:biovocabbert_constr} (left), we find that it achieves a 74\% segmentation F1 score on the biomedical SIGMORPHON subset, a very strong result compared to current tokenizers. For reference, the best segmentation F1 score reported in the English word-level test set of the SIGMORPHON 2022 Shared Task is 93.7\% by the DeepSpin team \cite{peters-martins-2022-beyond}. Even though this score is not comparable to ours due to our use of a biomedical development subset for evaluation, we note that our fine-tuned CANINE model's performance is quite strong given that it is designed for pure tokenization as explained above.

\subsection{BioVocabBERT: Domain Knowledge Injection}

Despite its satisfactory segmentation performance, it is challenging to use our CANINE-based segmentation model as a language model tokenizer due to its vocabulary-less nature. Since this model can segment words arbitrarily using our character classification framework, unseen words can be split into subwords which have never been seen by the LM. Thus, pre-training a language model using this tokenizer would require allowing the model to increase its vocabulary size without limit while training and using an out-of-vocabulary token when unseen tokens are encountered during inference. This would lead to a language model with an exceedingly large vocabulary size (which would increase the cost of pre-training significantly) and potentially limited generalization ability to unseen tokens. 

To tackle this problem, we introduce a tokenizer which uses the same left-to-right decoding algorithm used by BERT and PubMedBERT but replace its vocabulary with one designed for biomedical segmentation. In order to build a vocabulary which covers important biomedical tokens, we leverage the Unified Medical Language System, a medical knowledge base which contains approximately 15 million medical concept phrases. As shown in Figure \ref{fig:biovocabbert_constr} (center), we extract all single words from UMLS concept phrases and segment each of them using the CANINE-based tokenizer. This produces around \num{250000} unique subwords which we further reduce to \num{55580} by eliminating ones which only appear once in the CANINE segmented set of UMLS words. In order to avoid segmenting standard English words in unintuitive ways due to the higher proportion of biomedical subwords, we augment our \num{55580} biomedical subwords with the original BERT vocabulary as seen in Figure \ref{fig:biovocabbert_constr} (right). After removing duplicate tokens, we are left with a vocabulary of \num{80181} tokens. This carefully designed vocabulary enables our new tokenizer, BioVocabBERT, to obtain a segmentation score of 48.5\% on the SIGMORPHON biomedical subset as seen in \ref{tab:segmentation-eval}, outperforming the best scoring current wordpiece-based tokenizer PubMedBERT by almost 30 points.

\section{Experimental Setup}

\subsection{Biomedical Pre-training}

In order to discover whether morpheme segmentation performance has an effect on the biomedical language model pre-training process, we compare the downstream performance of two language models using tokenizers with very distinct segmentation performance but otherwise equivalent pre-training processes. The first model is pre-trained using the same tokenizer as PubMedBERT while the other one uses our designed BioVocabBERT tokenizer and is thus referred to by the same name.

As other BERT based architecture models, we pre-train these using the masked language modeling objective and choose standard token masking percentages used in previous work \cite{pubmedbert}. We use the easily accessible and readily pre-processed corpus used for BlueBERT \cite{peng-etal-2020-empirical} pre-training which contains around 4 billion words\footnote{\url{https://github.com/ncbi-nlp/BlueBERT/blob/master/README.md\#pubmed}}. For pre-training, we base our implementation on the work by \citet{izsak-etal-2021-train} to obtain the most efficient pre-training possible. We describe the data, optimization steps, batch size, hardware and other pre-training details used for both models in Table \ref{tab:pretrain_details} and Appendix \ref{pretrain_details}.  

\subsection{Tasks}

In order to determine how tokenization improvements affect the quality of biomedical concept representation in language models, we narrow our task selection to those which are more closely related to entity understanding instead of overall sentence understanding as is the case with relation extraction, sentence similarity or natural language inference. We select named entity recognition (NER) and entity linking (EL), also referred to as concept normalization, as the two biomedical NLP tasks which most closely meet this criterion. \\

\noindent \textbf{NER.} We focus on evaluating our models in the more standard fully supervised fine-tuning NER setting as is done in previous work \cite{Lee2019BioBERTAP, pubmedbert}. We run hyperparameter tuning on the development set, the search space used can be found in Appendix \ref{sec:appendix}. 

We also study our models' low-resource NER performance using 500 and 1000 examples. We also carry out hyperparameter selection which can be found in Appendix \ref{sec:appendix}. We report results on the development set only for our low-resource NER experiments.

\noindent \textbf{Entity Linking.} For entity linking, we evaluate our models' zero-shot performance as done by \citet{liu-etal-2021-self} which allows us to measure entity representation quality as directly as possible.

\subsection{Datasets}

For NER, we use all datasets from BLURB \cite{pubmedbert}, a comprehensive biomedical NLP benchmark. For entity linking, we follow previous work \cite{liu-etal-2021-self} and use four popular entity linking datasets, three of which are also included in BLURB as NER datasets. All dataset names and statistics can be found in Table \ref{tab:dataset_stats}. Below, we provide brief descriptions for each dataset we use, for more information about processing and training splits for these datasets,  we refer the interested reader to the dataset descriptions in \citet{pubmedbert} and \citet{liu-etal-2021-self}.\\

\begin{table}[!h]
\small
\centering
\resizebox{\columnwidth}{!}{%
\begin{tabular}{@{}lcccccc@{}}
\toprule
   & 
   \textbf{NER} &
   \textbf{EL} &
  \textbf{Train} &
  \textbf{Dev} &
  \textbf{Test} &
   \\
  \midrule
\textbf{BC5CDR-disease} &X&X   & \num{4182} & \num{4244} & \num{4424}\\
\textbf{BC5CDR-chem} &X&X& \num{5203} & \num{5347} & \num{5385}\\
\textbf{NCBI-Disease}   &X&X  &  \num{5134} & \num{787} & \num{960}\\
\textbf{JNLPBA}       &X&    &  \num{46750} & \num{4551} & \num{8662}\\
\textbf{BC2GM}        &X&    & \num{15197} & \num{3061} & \num{6325}\\ 
\textbf{MedMentions} &&X & \num{282091} & \num{71062} & \num{70405}\\
\bottomrule
\end{tabular}%
}
\caption{Dataset statistics.}

\label{tab:dataset_stats}
\end{table}

\noindent \textbf{BC5CDR.}
The BioCreative V Chemical-Disease Relation corpus \cite{jiaoli} contains both disease and chemical annotations on PubMed abstracts. We evaluate disease and chemical entity extraction and linking separately following previous work \cite{pubmedbert}.\\
\noindent \textbf{NCBI-Disease.}
The Natural Center for Biotechnology Information Disease corpus \cite{ncbi} contains disease name and concept annotations for \num{793} PubMed abstracts. \\
\noindent \textbf{JNLPBA.}
The Joint Workshop on Natural Language Processing in Biomedicine and its Applications dataset \cite{jnlpba} contains \num{2000} abstracts from MEDLINE selected and annotated by hand for gene related entities. \\
\noindent \textbf{BC2GM.}
The Biocreative II Gene Mention corpus \cite{bc2gm} contains \num{17500} sentences from PubMed abstracts labeled for gene entities.\\
\noindent \textbf{MedMentions.} MedMentions \cite{Mohan2019MedMentionsAL} is a large-scale entity linking dataset containing over 4,000 abstracts and around 350,000 mentions linked to the 2017AA version of UMLS.

\section{Results \& Discussion}

\subsection{Fully-Supervised NER}

As seen in Table \ref{tab:ner_results}, our language models obtain competitive fully-supervised NER results compared to the results reported by \citet{pubmedbert}, validating our pre-training and fine-tuning process. We first find that the differences in performance between our PubMedBERT and BioVocabBERT models are very small and inconsistent across NER datasets. We note that the difference in performance between these models is often within the standard deviation reported within each dataset. Additionally, we see no pattern in performance differences based on entity types. For disease NER, BioVocab underperforms on NCBI-Disease but overperforms in BC5CDR-disease while in gene based NER, BioVocabBERT outperforms by a slightly larger margin on JNLPBA but performs only on-par on BC2GM. This seems to suggest that, at least when fine-tuning on a significant number of examples, PubMedBERT can very adequately compensate for instances of sub-optimal biomedical segmentation.   

\begin{table}[!h]
\centering
\resizebox{\columnwidth}{!}{%
\begin{tabular}{@{}lccc@{}}
\toprule
                        & \textbf{PubMedBERT$^{*}$} & \textbf{PubMedBERT} & \textbf{BioVocabBERT} \\ \midrule
\textbf{NCBI-Disease}   & 87.8 & 87.1 $\pm$ 0.8                & 86.7 $\pm$ 0.4               \\
\textbf{BC5CDR-disease} & 85.6 & 84.7 $\pm$ 0.2 & 85.2 $\pm$ 0.3            \\
\textbf{BC5CDR-chem}    & 93.3 & 93.0 $\pm$ 0.3     & 93.4  $\pm$ 0.4              \\
\textbf{JNLPBA}         & 79.1 & 78.2 $\pm$ 0.6      & 78.9 $\pm$ 0.1               \\
\textbf{BC2GM}          & 84.5 & 83.4 $\pm$ 0.2       & 83.5  $\pm$ 0.3             \\\bottomrule
\end{tabular}%
}
\caption{Comparison of fully supervised NER performance for the originally reported PubMedBERT, denoted by $^{*}$, our PubMedBERT replica and our BioVocabBERT model. We report 3 runs for each of our models and provide the average entity-level F1 on the test set along with its standard deviation.}
\label{tab:ner_results}
\end{table}

\subsection{Low-Resource NER}

To explore whether parity in fully-supervised NER comes from the effects of large scale fine-tuning or from the underlying models' entity representation quality, we carry out a low-resource NER study using only \num{500} and \num{1000} training examples. We present results only on the development set for this setting. Our results suggest that, even when fewer training examples are used for fine-tuning, the difference between models is small. As shown in Table \ref{tab:low-res-ner}, BioVocabBERT obtains small and inconsistent improvements in downstream performance over PubMedBERT tokenization across NER datasets in the low-resource setting. Nevertheless, we note that the largest gains for BioVocabBERT in these low data regimes come from chemical and genetic NER datasets (BC2GM and JNLPBA), suggesting that our tokenization strategy could be especially beneficial for irregular genetic entities.

\begin{table}[!h]
\small
\centering
\resizebox{\columnwidth}{!}{%
\begin{tabular}{@{}lcccccc@{}}
\toprule
\textbf{}                     & \multicolumn{2}{c}{\textbf{PubMedBERT}} & \multicolumn{2}{c}{\textbf{BioVocabBERT}} & \multicolumn{2}{c}{\textbf{Percent $\Delta$}}\\ \midrule
\multicolumn{1}{r}{\textbf{}} & \textbf{500}        & \textbf{1000}       & \textbf{500}        & \textbf{1000} & \textbf{500}        & \textbf{1000}       \\ \midrule
\textbf{NCBI-disease}         &$ 77.2                $&$ 81.2                $&$ 77.7                $&$ 80.6 $&$ 0.5 $&$ -0.6 $     \\
\textbf{BC5CDR-disease}       &$ 79.0                $&$ 81.4                $&$ 79.3                $&$ 81.6 $&$ 0.3 $&$ 0.2$\\
\textbf{BC5CDR-chem}          &$ 91.7                $&$ 92.2                $&$ 92.1                $&$ 92.8             $&$ 0.4 $&$ 0.6$   \\
\textbf{BC2GM}                &$ 69.5                $&$ 75.5                $&$ 71.5                $&$ 76.9 $&$ 2.0 $&$ 1.4$\\
\textbf{JNLPBA}               &$ 75.6                $&$ 77.2                $&$ 76.3                $&$ 77.6             $&$ 0.7 $&$ 0.4 $  \\ \bottomrule
\end{tabular}%
}
\caption{Comparing our models on low-resource NER with \num{500} and \num{1000} examples. We report the entity-level F1 on the development set for this setting.}
\label{tab:low-res-ner}
\end{table}

\begin{table*}[!t]
\centering
\resizebox{\textwidth}{!}{%
\begin{tabular}{@{}lll@{}}
\toprule
\textbf{Sub-optimal Tokenization}                                                                              & \textbf{Word Embedding 5-NN}                               & \textbf{{[}CLS{]} Embedding 5-NN}                           \\ \midrule
\multirow{5}{*}{\begin{tabular}[c]{@{}l@{}}epicarditis\\ (epic-ardi-tis)\end{tabular}}                       & epicardiectomy  (epic-ardi-ectomy)                         & pancarditis  (panc-ardi-tis)                                \\
                                                                                                             & pancarditis  (panc-ardi-tis)                               & \textbf{perimyocarditis  (peri-my-ocardi-tis)}              \\
                                                                                                             & epicardin  (epic-ardi-n)                                   & \textbf{myopericarditis  (myo-peri-car-di-tis)}             \\
                                                                                                             & epicardium  (epic-ardi-um)                                 & \textbf{myoendocarditis  (myo-end-ocardi-tis)}              \\
                                                                                                             & endopericarditis  (endop-eric-ardi-tis)                    & pleuropericarditis  (pleu-rop-eric-ardi-tis)                \\  \midrule
\multirow{5}{*}{\begin{tabular}[c]{@{}l@{}}neuromodulation\\ (neuromod-ulation)\end{tabular}}                & neuromodulations  (neuromod-ulations)                      & neuromodulations  (neuromod-ulations)                       \\
                                                                                                             & neuromodulators  (neuromod-ulators)                        & neuromodulators  (neuromod-ulators)                         \\
                                                                                                             & neuromodulator  (neuromod-ulator)                          & neuromodulator  (neuromod-ulator)                           \\
                                                                                                             & immunomodulation  (immunomod-ulation)                      & \textbf{neuroexcitation  (neuro-exc-itation)}               \\
                                                                                                             & immunoregulation  (immunoreg-ulation)                      & \textbf{neuroregulation  (neuro-reg-ulation)}               \\ \bottomrule
\end{tabular}%
}
\caption{In this table we show two PubMedBERT sub-optimal tokenization examples and their nearest neighbors with respect to word embeddings and `[CLS]' token embeddings. Neighbors in \text{bold} are terms that were missed by the word embeddings but are retrieved correctly by the `[CLS]' embeddings, repairing the sub-optimal tokenization bias.}
\label{tab:robustness_analysis}
\end{table*}

\subsection{Zero-Shot Entity Linking}

Following our low-resource NER results, we evaluate entity representation quality even more directly by measuring the zero-shot entity linking performance of both models. As shown in Table \ref{tab:entity_link}, performance of our models exceeds the original PubMedBERT results reported by \citet{liu-etal-2021-self}, validating the quality of our pre-training procedure. We note that the main difference between our pre-training setup and the original PubMedBERT setting is the use of the masked language modeling (MLM) objective alone instead of both MLM and next-sentence prediction (NSP) objectives. This suggests that the use of the MLM objective only might be better aligned with obtaining high quality entity representations.

\begin{table}[!h]
\centering
\resizebox{\columnwidth}{!}{%
\begin{tabular}{@{}lcccccc@{}}
\toprule
\textbf{}               & \multicolumn{2}{c}{\textbf{PubMedBERT$^{*}$}} & \multicolumn{2}{c}{\textbf{PubMedBERT}} & \multicolumn{2}{c}{\textbf{BioVocabBERT}} \\ \midrule
\textbf{}               & \textbf{R@1}                & \textbf{R@5}                & \textbf{R@1}                 & \textbf{R@5}                 & \textbf{R@1}                 & \textbf{R@5}                 \\ \midrule
\textbf{NCBI-Disease}   & 77.8             & 86.9             & 88.5              & 93.5              & 87.6              & 92.0              \\
\textbf{BC5CDR-disease} & 89.0             & 93.8             & 91.7              & 95.0              & 91.0              & 94.1              \\
\textbf{BC5CDR-chem}    & 93.0             & 94.6             & 95.3              & 96.1              & 95.4              & 95.9              \\
\textbf{MedMentions}    & 43.9             & 64.7             & 44.9              & 65.4              & 45.4              & 65.9              \\ \bottomrule
\end{tabular}%
}
\caption{Comparison of zero-shot entity linking performance for the originally reported PubMedBERT, denoted by $^{*}$, our PubMedBERT replica and our BioVocabBERT model.}
\label{tab:entity_link}
\end{table}

Additionally, when comparing our models, we find that BioVocabBERT slightly underperforms the PubMedBERT replica on all datasets except the more diverse MedMentions dataset. However, we note that the improvements obtained in the MedMentions dataset are also quite small at under 1\%. Given the zero-shot nature of this experiment, it suggests that the entity representations obtained by these two models are of comparable quality and that PubMedBERT's pre-training enables a high degree of robustness around sub-optimal tokenization.  

\subsection{Case Study: Tokenization Robustness}
\label{case_study}

As shown in the NER and entity linking experiments above, the downstream performance of biomedical language models appears to be mostly robust to  biomedical concept segmentation which is not well-aligned with human judgments. To analyze this phenomenon, we take a closer look at how our pre-trained PubMedBERT model represents biomedical concepts which are segmented in apparently erroneous ways by the PubMedBERT tokenizer. Table \ref{tab:robustness_analysis} contains two words from the biomedical subset of SIGMORPHON which were sub-optimally segmented by PubMedBERT. For each of these words we include two sets of their 5 nearest neighbors according to different embedding types. The first set shows the nearest neighbors obtained using embeddings computed by averaging all subword embeddings that make up a specific word. The second set comes from using the `[CLS]' token embedding of our PubMedBERT model, often used for downstream tasks in standard fine-tuning.  The pool of words from which these neighbors are obtained consists of all the unique UMLS words in UMLS phrases used in the construction of BioVocabBERT. 

Since it comes directly from subword embeddings, the first set of neighbors is meant to show whether the tokenizer's sub-optimal segmentation introduces a bias which distracts the model from the true semantics of a biomedical term. The second neighborhood is meant to more faithfully show us how the model represents a biomedical concept. Comparing these two sets can let us determine if the bias introduced by subword embeddings is successfully regulated by the overall model. 

We first observe that the bias we expected to find in the word embedding neighborhoods is evidently present. Most words in these first sets are segmented in exactly the same ways as the original sub-optimally segmented word. As seen in Table \ref{tab:robustness_analysis}, the word `neuromodulation' is segmented by PubMedBERT as `neuromod-ulation', splitting the meaningful 'modulate' morpheme down the middle, an example of the compound error in Table \ref{tab:seg_error_types}. Due to this, other words with the same subword but different semantics such as `immunomodulation' (`immunomod-ulation') and `immunoregulation' (`immonoreg-ulation') are added to the word embedding neighborhood. This is also seen in the second example, where the word embedding neighbors of `epicarditis' (`epic-ardi-tis') all contain at least two of the three original subwords. If these word embeddings were the final model representations, this bias could lead to considerable errors in downstream tasks like entity linking by up-weighting terms based on sub-optimal subwords. 

Fortunately, we observe that the language model is able to readily overcome the bias observed in the word embeddings when it comes to the final `[CLS]' representations. The second neighborhoods often contain semantically relevant words which were segmented differently than the original, such as `neuroexcitation' (`neuro-exc-itation') and `neuroregulation' (`neuro-reg-ulation') for `neuromodulation' (`neuromod-ulation') or `perimyocarditis' (`peri-my-ocardi-tis') for `epicarditis' (`epic-ardi-tis'), which both mean types of inflammation of the pericardium. This shows us that the language model successfully extracts the semantics of the morpheme `neuro' from `neuromod' as well as the cardiovascular related semantics from both `epic-ardi' and `ocardi', effectively mitigating the detrimental effects seen in the word embedding neighborhoods from sub-optimal tokenization. We thus conclude that this same robustness is responsible for the parity observed in downstream tasks between BioVocabBERT and the original PubMedBERT. More examples which show similar trends as the ones in Table \ref{tab:robustness_analysis} can be found in Table \ref{tab:robustness_analysis_complete} in Appendix \ref{sec:neighborhoods}. 

\subsection{Word Frequency Study}

\begin{figure}[!t]
     \centering
     \includegraphics[width=\columnwidth]{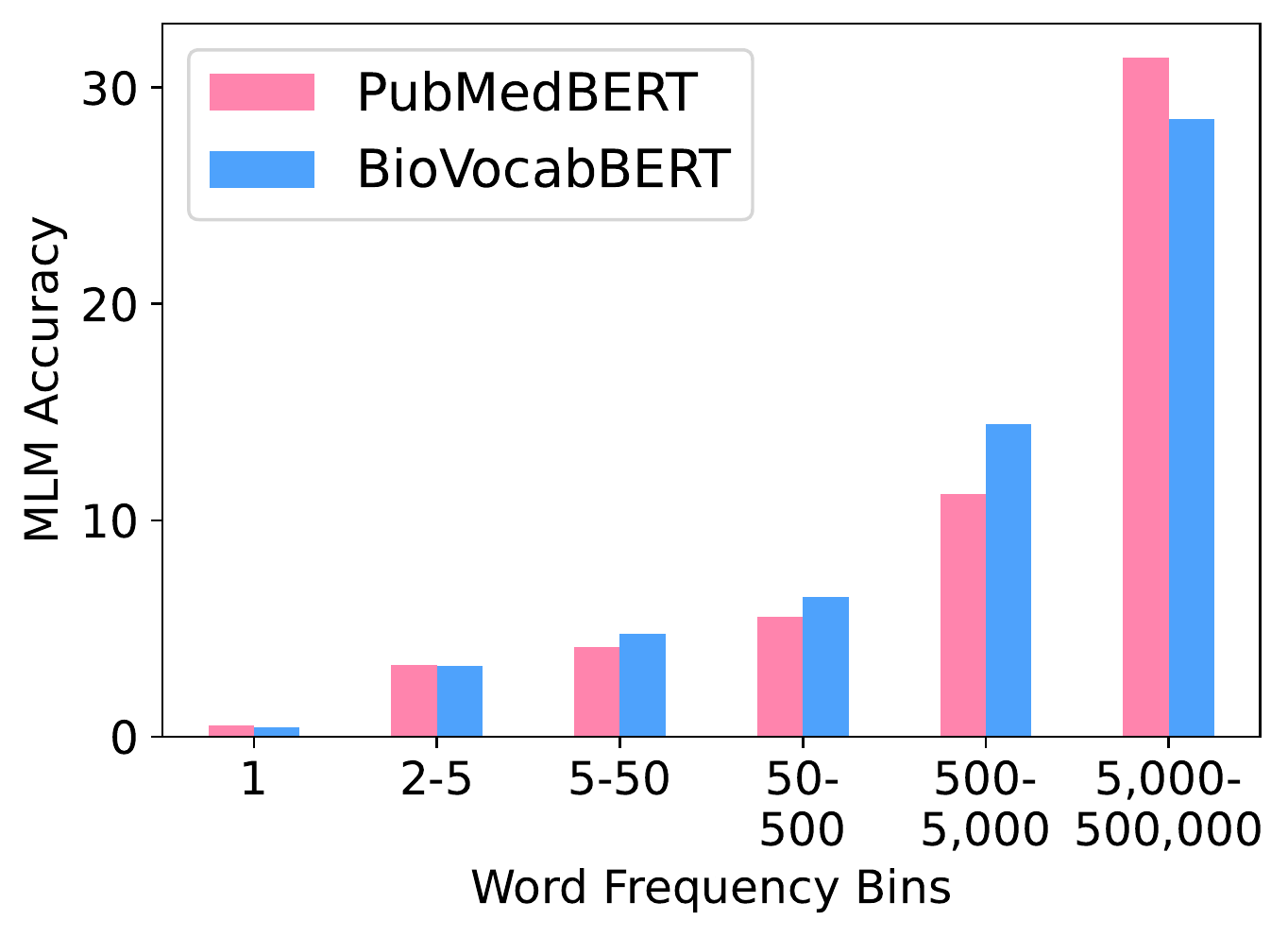}
     \caption{MLM accuracy for our pre-trained models averaged across 10,000 word instances which fall under each word frequency bin.}
     \label{fig:mlm_acc}
     \vspace{-5pt}
\end{figure}

All the findings above suggest that biomedical language model pre-training yields entity representations which are fairly robust to tokenization failures. However, it is important to note that the distribution of rare vs. frequent entities in these small and medium scale datasets will be naturally skewed towards frequent entities if not intentionally manipulated. Therefore, we design an experiment which explores whether the quality of representations in BioVocabBERT and our PubMedBERT replica varies with respect to word frequency. In this experiment, we obtain \num{10000} instances of words from the pre-training corpus in each of the frequency bins listed in Table \ref{fig:mlm_acc}. We encode the sentence in which each instance is found and mask the word of interest. We report the percentage of words which are predicted correctly using the masked language modeling head's prediction in each bin. 

We note that both models perform very similarly across frequency bins until the two bins with the largest frequencies. Our BioVocabBERT model obtains a somewhat significant advantage in the second highest \num{500}-\num{5000} frequency bin which then inverts to a similarly significant drop in the category with the most frequent words. This trade-off is likely due to many high frequency words having a single token in the PubMedBERT vocabulary, which leads the model to have a natural bias towards predicting these words. In medium frequency words, PubMedBERT's bias towards high frequency words is likely detrimental and BioVocabBERT is able to easily outperform it. This trade-off appears to be small enough to have little effect on downstream performance for these models but we leave exploring its effect for future work.

\section{Conclusion}

In this work, we first note that current biomedical tokenization methods are not well aligned with human judgments and highlight that the agglutinating nature of biomedical terminology could be affected by this sub-optimal segmentation. To understand whether this limited segmentation performance has an effect on downstream applications, we first build a biomedical tokenizer which is better aligned with human judgments using supervised morpheme segmentation and biomedical domain-knowledge. We then leverage this tokenizer to pre-train our BioVocabBERT model and compare it with a replicated PubMedBERT model on the NER and entity linking tasks. Surprisingly, we find that these models achieve almost exact parity in all datasets evaluated, suggesting that PubMedBERT's domain specific tokenization and pre-training process was already quite robust to sub-optimal tokenization. We further verify this idea with a case-study which qualitatively confirms our observations. We hope that our work can give researchers and practitioners some insight into how instances of sub-optimal segmentation, which are often jarring to human experts, could have little effect on a model's downstream applicability.

\section{Limitations}

Although our findings suggest that biomedical language model pre-training is quite robust to sub-optimal tokenization, we note that our work has a few potential limitations that should be explored further. The use of a biomedically relevant subset of the SIGMORPHON Shared Task dataset for evaluating biomedical term tokenization is a straight-forward and reasonable strategy, however, it is important to highlight that the resource was not created for this purpose and might not be perfectly aligned with ideal biomedical tokenization. Additionally, we would like to point out that even though our BioVocabBERT tokenizer outperforms other equivalent tokenizers like PubMedBERT's, it severly underperforms the best possible segmentation accuracy (48.5 vs 74.1 for our fine-tuned CANINE model). It is therefore possible,  although unexpected, that a tokenizer which performs biomedical tokenization at even higher levels could lead to sudden improvements in the pre-training process. Finally, we note that the effects of the BioVocabBERT's much larger vocabulary size, almost three times larger than PubMedBERT's, on the pre-training process were not explored in depth. Nevertheless, given that some previous work \cite{Feng2022PretrainingWW} argues that larger vocabularies lead to slight improvements in downstream tasks, our main conclusions are likely to hold.

\section*{Acknowledgements}
The authors would like to thank the anonymous reviewers and colleagues from the
OSU NLP group for their valuable feedback. 
This research was supported in part by NIH R01LM014199 and Ohio Supercomputer Center \citep{OhioSupercomputerCenter1987}.

\bibliography{anthology,custom}
\bibliographystyle{acl_natbib}

\clearpage
\appendix

\section{NER Hyperparameter Tuning}
\label{sec:appendix}

We run hyperparameter tuning for each model in the fully-supervised and low-resource NER settings and they can both be found in Tables \ref{tab:hyperparameter_search_grid} and \ref{tab:hyperparameter_search_grid_low_res} below.

\begin{table}[!h]
\centering
\small
\resizebox{\columnwidth}{!}{%
\begin{tabular}{@{}lccccc@{}}
\toprule
 & \multicolumn{1}{c}{\textbf{\begin{tabular}[c]{@{}c@{}}Learning \\Rate\end{tabular}}}                       & \multicolumn{1}{c}{\textbf{\begin{tabular}[c]{@{}c@{}}Batch\\ Size\end{tabular}}}        & \multicolumn{1}{c}{\textbf{\begin{tabular}[c]{@{}c@{}}Warmup \\Ratio\end{tabular}}}         & \multicolumn{1}{c}{\textbf{\begin{tabular}[c]{@{}c@{}}Weight \\Decay\end{tabular}}}               & \multicolumn{1}{c}{\textbf{\begin{tabular}[c]{@{}c@{}}Total\\ Epoch \\Number\end{tabular}}} \\ \midrule
\textbf{\begin{tabular}[c]{@{}l@{}}Search\\ Space\end{tabular}}  & \begin{tabular}[c]{@{}c@{}}\num{1}e-\num{5}\\ \num{3}e-\num{5}\end{tabular} & \begin{tabular}[c]{@{}c@{}}\num{16}\\ \num{32}\end{tabular} & \begin{tabular}[c]{@{}c@{}}\num{0.06}\end{tabular} & \begin{tabular}[c]{@{}c@{}}\num{0.1}\end{tabular} & \begin{tabular}[c]{@{}c@{}}\num{5}\\ \num{10}\end{tabular}                       \\ \bottomrule
\end{tabular}%
}
\caption{Hyperparameter search grid used for fully-supervised NER experiments.}
\label{tab:hyperparameter_search_grid}
\end{table}

\begin{table}[!h]
\centering
\small
\resizebox{\columnwidth}{!}{%
\begin{tabular}{@{}lccccc@{}}
\toprule
 & \multicolumn{1}{c}{\textbf{\begin{tabular}[c]{@{}c@{}}Learning \\Rate\end{tabular}}}                       & \multicolumn{1}{c}{\textbf{\begin{tabular}[c]{@{}c@{}}Batch\\ Size\end{tabular}}}        & \multicolumn{1}{c}{\textbf{\begin{tabular}[c]{@{}c@{}}Warmup \\Ratio\end{tabular}}}         & \multicolumn{1}{c}{\textbf{\begin{tabular}[c]{@{}c@{}}Weight \\Decay\end{tabular}}}               & \multicolumn{1}{c}{\textbf{\begin{tabular}[c]{@{}c@{}}Total\\ Epoch \\Number\end{tabular}}} \\ \midrule
\textbf{\begin{tabular}[c]{@{}l@{}}Search\\ Space\end{tabular}}  & \begin{tabular}[c]{@{}c@{}}\num{1}e-\num{5}\\ \num{3}e-\num{5}\end{tabular} & \begin{tabular}[c]{@{}c@{}}\num{16}\\ \num{32}\end{tabular} & \begin{tabular}[c]{@{}c@{}}\num{0.06}\end{tabular} & \begin{tabular}[c]{@{}c@{}}\num{0.1}\end{tabular} & \begin{tabular}[c]{@{}c@{}}\num{15}\\ \num{25}\end{tabular}                                    \\ \bottomrule
\end{tabular}%
}
\caption{Hyperparameter search grid used for low-resource NER experiments.}
\label{tab:hyperparameter_search_grid_low_res}
\end{table}

\section{Pre-training Details}
\label{pretrain_details}

Our models were pre-trained on 4 80GB A100s. The process took approximately 2 and 3 days respectively for PubMedBERT and BioVocabBERT given the larger computational requirements of using an \num{80000} subword vocabulary. 

\begin{table}[!h]
\centering
\resizebox{\columnwidth}{!}{%
\begin{tabular}{@{}lcccccc@{}}
\toprule
&\textbf{Objectives}&\textbf{\begin{tabular}[c]{@{}c@{}}Vocab.\\ Size\end{tabular}}& \textbf{\begin{tabular}[c]{@{}c@{}}Corpus\\ Size\end{tabular}}     & \textbf{\begin{tabular}[c]{@{}c@{}}Gradient\\ Steps\end{tabular}} & \textbf{\begin{tabular}[c]{@{}c@{}}Batch\\ Size\end{tabular}} & \textbf{\begin{tabular}[c]{@{}c@{}}\# of \\Examples\end{tabular}} \\ \midrule
\textbf{\begin{tabular}[c]{@{}l@{}}PubMedBERT\\ (Original)\end{tabular}}  & \begin{tabular}[c]{@{}c@{}}MLM \\\& NSP\end{tabular} & 28,895  & \begin{tabular}[c]{@{}c@{}}21GB\end{tabular} & 62,500                   & 8,192               & 512M                        \\\midrule
\textbf{\begin{tabular}[c]{@{}l@{}}PubMedBERT \\(Replica)\end{tabular}} &MLM & 28,895 & 25GB                     & 62,500                   & 8,192               & 512M                        \\\midrule
\textbf{\begin{tabular}[c]{@{}l@{}}BioVocabBERT\end{tabular}} &MLM & 80,181  & 25GB                     & 62,500                   & 8,192               & 512M                        \\ \bottomrule
\end{tabular}%
}
\caption{Pre-training details for the original PubMedBERT compared to our models.}
\label{tab:pretrain_details}
\end{table}

\section{More Neighborhood Examples}
\label{sec:neighborhoods}

The neighborhood examples shown in Table \ref{tab:robustness_analysis_complete}, help demonstrate that the general trends discussed in \S\ref{case_study} hold more generally for many sub-optimal segmentation examples.  

\begin{table*}[]
\centering
\resizebox{\textwidth}{!}{%
\begin{tabular}{@{}lll@{}}
\toprule
\textbf{Sub-optimal Segmentation}                                                                              & \textbf{Word Embedding 5-NN}                               & \textbf{{[}CLS{]} Embedding 5-NN}                           \\ \midrule
\multirow{5}{*}{\begin{tabular}[c]{@{}l@{}}abdominopelvic\\ (abdom-ino-pe-lv-ic)\end{tabular}}               & abdominopelvis  (abdom-ino-pe-lv-is)                       & abdominocentesis  (abdom-ino-cent-esis)                     \\
                                                                                                             & sacropelvic  (sacro-pe-lv-ic)                              & thoracopelvic  (thorac-ope-lv-ic)                           \\
                                                                                                             & uteropelvic  (utero-pe-lv-ic)                              & midpelvic  (mid-pe-lv-ic)                                   \\
                                                                                                             & abdomino  (abdom-ino)                                      & sacropelvic  (sacro-pe-lv-ic)                               \\
                                                                                                             & midpelvic  (mid-pe-lv-ic)                                  & \textbf{extrapelvic  (extrap-elvic)}                        \\ \midrule
\multirow{5}{*}{\begin{tabular}[c]{@{}l@{}}neuroradiography\\ (neuroradi-ography)\end{tabular}}              & roentgenography  (roentgen-ography)                        & neuroradiology  (neuroradi-ology)                           \\
                                                                                                             & ventriculography  (ventricul-ography)                      & neuroradiologic  (neuroradi-ologic)                         \\
                                                                                                             & neuroradiology  (neuroradi-ology)                          & encephalography  (encephal-ography)                         \\
                                                                                                             & electroretinography  (electroretin-ography)                & \textbf{neurography  (neuro-graphy)}                        \\
                                                                                                             & herniography  (herni-ography)                              & cerebroangiography  (cerebro-angi-ography)                  \\ \midrule
\multirow{5}{*}{\begin{tabular}[c]{@{}l@{}}postinfectional\\ (postin-fection-al)\end{tabular}}               & postinfection  (postin-fection)                            & reinfection  (rein-fection)                                 \\
                                                                                                             & postin  (postin)                                           & postinfection  (postin-fection)                             \\
                                                                                                             & postinjection  (postin-jection)                            & \textbf{superinfection  (super-infection)}                  \\
                                                                                                             & reinfection  (rein-fection)                                & reinfected  (rein-fected)                                   \\
                                                                                                             & postinfusion  (postin-fusion)                              & \textbf{superinfections  (super-infection-s)}               \\ \midrule
\multirow{5}{*}{\begin{tabular}[c]{@{}l@{}}neurogastrointestinal\\ (neuro-ga-st-ro-intestinal)\end{tabular}} & extragastrointestinal  (extra-ga-st-ro-intestinal)         & extragastrointestinal  (extra-ga-st-ro-intestinal)          \\
                                                                                                             & pangastrointestinal  (pan-ga-st-ro-intestinal)             & pangastrointestinal  (pan-ga-st-ro-intestinal)              \\
                                                                                                             & myoneurogastrointestinal  (myo-ne-uro-ga-st-ro-intestinal) & enteropancreatic  (enter-opancre-atic)                      \\
                                                                                                             & gastrogastric  (gastro-ga-st-ric)                          & \textbf{gastroenteropancreatic  (gastroenter-opancre-atic)} \\
                                                                                                             & gastrogastrostomy  (gastro-ga-st-rost-omy)                 & nasopancreatic  (nas-opancre-atic)                          \\ \midrule
\multirow{5}{*}{\begin{tabular}[c]{@{}l@{}}adrenocorticosteroid\\ (adrenocortic-oster-oid)\end{tabular}}     & adrenocorticosteroids  (adrenocortic-oster-oids)           & adrenocorticosteroids  (adrenocortic-oster-oids)            \\
                                                                                                             & glucocorticosteroid  (glucocortic-oster-oid)               & adrenocorticotropic  (adrenocortic-otropic)                 \\
                                                                                                             & mineralocorticosteroid  (mineralocortic-oster-oid)         & \textbf{corticoids  (cortic-oids)}                          \\
                                                                                                             & mineralocorticosteroids  (mineralocortic-oster-oids)       & corticoid  (cortic-oid)                                     \\
                                                                                                             & glucosteroid  (gluc-oster-oid)                             & glucosteroid  (gluc-oster-oid)                              \\ \bottomrule
\end{tabular}%
}
\caption{In this table we show more PubMedBERT sub-optimal segmentation examples and their nearest neighbors with respect to word embeddings and `[CLS]' token embeddings. As in Table \ref{tab:robustness_analysis}, \text{bold} neighbors were missed by the word embeddings but are retrieved correctly by the `[CLS]' embeddings, repairing the sub-optimal segmentation bias.}
\label{tab:robustness_analysis_complete}
\end{table*}

\end{document}